\documentclass{article}

\PassOptionsToPackage{numbers}{natbib}

\usepackage[final]{neurips_2021}

\usepackage[utf8]{inputenc} %
\usepackage[T1]{fontenc}    %
\usepackage{hyperref}       %
\usepackage{url}            %
\usepackage{booktabs}       %
\usepackage{amsfonts}       %
\usepackage{nicefrac}       %
\usepackage{microtype}      %
\usepackage{xcolor}         %
\usepackage{graphicx}
\usepackage{float}
\usepackage{wrapfig}
\usepackage{comment}
\usepackage{caption}
\usepackage{subcaption}

\title{Two-phase training mitigates class imbalance for camera trap image classification with CNNs}

\author{%
  Farjad Malik \\
  \And
  Simon Wouters\thanks{Joint first author.} \\
   \And
   Ruben Cartuyvels\thanks{Corresponding author.} \\
  \texttt{ruben.cartuyvels@kuleuven.be} \\
   \And
  Erfan Ghadery \\
 \texttt{erfan.ghadery@kuleuven.be} \\
 \And
  Marie-Francine Moens \\
  \texttt{sien.moens@kuleuven.be}
 \And
  {\normalfont Department of Computer Science} \\
  KU Leuven \\
}

\begin{document}

\maketitle

\begin{abstract}

By leveraging deep learning to automatically classify camera trap images, ecologists can monitor biodiversity conservation efforts and the effects of climate change on ecosystems more efficiently.  
Due to the imbalanced class-distribution of camera trap datasets, current models are biased towards the majority classes. 
As a result, they obtain good performance for a few majority classes but poor performance for many minority classes.
We used two-phase training to increase the performance for these minority classes. 
We trained, next to a baseline model, four models that implemented a different versions of two-phase training on a subset of the highly imbalanced Snapshot Serengeti dataset. 
Our results suggest that two-phase training can improve performance for many minority classes, with limited loss in performance for the other classes. 
We find that two-phase training based on majority undersampling increases class-specific F1-scores up to 3.0\%. 
We also find that two-phase training outperforms using only oversampling or undersampling by 6.1\% in F1-score on average. Finally, we find that a combination of over- and undersampling leads to a better performance than using them individually.
  
\end{abstract}
\section{Introduction}
A recent report by the World Wide Fund for Nature (WWF) confirms that biodiversity and ecosystems are deteriorating worldwide \citep{almond2020living}. 
Population sizes of mammals, birds, amphibians, reptiles and fish have decreased by an average of 68\% between 1970 and 2016 across the world. 
This decrease in biodiversity has several causes, such as habitat loss due to pollution, species overexploitation 
or climate change.
Biodiversity
is important since it is a key indicator of overall healthy ecosystems which in their turn have important social and economic consequences for humans.
In particular, biodiversity and ecosystems influence our water quality, 
air quality and climate, they secure our food production and impact the spread of infectious diseases originating from animals \citep{almond2020living,diaz2019summary}.

Machine learning (ML) can help to more efficiently measure and monitor the well-being of ecosystems and the success of biodiversity conservation efforts \citep{huynh2018annotation,joly2020lifeclef,van2014nature,park2020illuminating}.
As an example, this paper proposes a method for automatic classification of camera trap images, a type of motion triggered cameras used in biological studies to estimate animal population density and activity patterns \citep{ridout2009estimating, foster2012critique,rowcliffe2014quantifying,sollmann2018gentle,tabak2019machine, trolliet2014use}.

Since manually labeling large numbers of camera trap images is time consuming and costly \citep{kelly2008estimating},
ML could be used to automatically detect animals and the species to which they belong in images. 
This work uses Convolutional Neural Networks 
\citep{lecun1998gradient,lecun2015deep}
to classify camera trap images. 
Training a CNN on a dataset of camera trap images is challenging, because camera trap images often only depict a part of an animal, because of high intra-class variation
due to differences in backgrounds, and because the class-distribution of camera trap datasets is typically highly imbalanced. 
This imbalance is inherent to camera trap datasets since it reflects the imbalance of ecosystems \citep{trebilco2013ecosystem}, and it results in biased classifiers that perform very well for a few majority classes but poorly for many minority classes. 
Classifiers that perform well on all classes would be of more value to ecologists, and moreover, rare or less observed animal species might even be of special interest to research.
Therefore, solutions are needed to mitigate this imbalance when 
classifying camera trap images. 

To this end, we use a two-phase training method \citep{lee2016plankton} to mitigate class imbalance, for the first time to the best of our knowledge on camera trap images. In experiments we compare it to different data-level class imbalance mitigation techniques, and show that it improves performance on minority classes, with limited loss in performance for other classes, resulting in an increase in macro F1-score.

\section{Related work}

Pioneering studies 
that automatically classified camera trap images relied on manual feature extraction and smaller datasets \citep{figueroa2014fast,swinnen2014novel,yu2013automated,chen2014deep}. Better and more scalable results were later achieved with deep CNNs and larger datasets \citep{gomez2016towards,norouzzadeh2018automatically,tabak2019machine, willi2019identifying, schneider2020three}.
Generally, models trained by these scholars achieve accuracies well above 90\%, but the models are biased towards majority classes, which severely affects their class-specific performance. 
Especially the 
performance for rare species is poor.
Scholars dealt with this challenge by removing the rare classes from the dataset \citep{gomez2016towards,willi2019identifying}, with confidence thresholding and letting experts review the uncertain classifications \citep{willi2019identifying}, with weighted losses, oversampling and emphasis sampling \citep{norouzzadeh2018automatically} or by using a combination of additional image augmentations for rare classes and novel sampling techniques \citep{schneider2020three}. 
Although \citep{norouzzadeh2018automatically} managed to greatly increase the accuracy for a few rare classes using oversampling, none of the aforementioned techniques systematically improved accuracy for most of the rare species. It can thus be concluded that dealing with class-imbalance in the context of camera trap image classification is still an unsolved issue.

Two categories of methods for mitigation of class imbalance in deep learning exist: data-level and algorithm-level techniques \citep{buda2018systematic, johnson2019survey}. 
The former refers to techniques that alter the class-distribution of the data, such as random minority oversampling (ROS) and random majority undersampling (RUS), 
which respectively randomly duplicate or randomly remove samples to obtain a more balanced dataset. 
More advanced techniques 
can also be used to synthesize new samples \citep{chawla2002smote,han2005borderline,he2008adasyn,wan2017variational,wang2017cgan,li2021plankton},
but these are computationally expensive, and they require a large number of images per class and images within a class that are sufficiently similar.
Algorithm-level techniques are techniques that work on the model itself, such as loss functions or thresholding \citep{lin2017focal,nemoto2018classification,buda2018systematic,johnson2019survey,he2013imbalanced,buda2018systematic}. 
Two-phase training, a hybrid technique, was recently introduced 
and shown to obtain good results 
for training a CNN classifier on a highly imbalanced dataset of images of plankton \citep{lee2016plankton}, and it was later used by others for image segmentation and classification \citep{havaei2017brain,buda2018systematic}. 
Because of these promising results and the broad applicability of 2-phase training, we 
test 2-phase training for camera trap images.

\section{Two-phase training}

Two-phase training consists of the following steps \citep{lee2016plankton}. $\mathcal{D}_{orig}$ is the original, imbalanced dataset. Figure \ref{fig:two-phase fig} in the appendix shows an overview of two-phrase training.
\begin{enumerate}
    \item \textbf{Phase 1}: a CNN $f_\theta$ is trained on a more balanced dataset $\mathcal{D}_{bal}$, obtained by any sampling method such as ROS, RUS or a combination thereof.
    \item \textbf{Phase 2}: the convolutional weights\footnote{I.e. all weights except the weights of the fully connected layers that project the CNN features to the classes.} of $f_\theta$ are frozen, and the network is trained further on the full imbalanced dataset $\mathcal{D}_{org}$.
\end{enumerate}
The 1st phase trains the convolutional layers with (more) equal importance allocated to minority classes, so they learn to extract relevant features for these classes as well. 
In the 2nd phase the classification layers learn to model the class imbalance present in the dataset.

\section{Dataset \& Experiments}
We used the 9th season of the publicly available Snapshot Serengeti (SS) dataset, 
which is generated by a grid of 225 cameras spread over the Serengeti National Park in Tanzania \citep{swanson2015snapshot}. 
The images were labeled by citizen scientists on the Zooniverse platform. 
After filtering some images, the full dataset $\mathcal{D}_{orig}$ contains 194k images belonging to 52 classes. 
The class-distribution of this dataset is depicted in fig. \ref{fig:season9_species fig} in the appendix, and is highly imbalanced, with the three majority classes accounting for just under 75\% of the data. 
We used this smaller subset of the full SS dataset for computational tractability, and to ensure insights remain valid for ecologists with access to smaller datasets.

Appendix \ref{app:hyperparams} lists the hyperparameters\footnote{Our code is publicly available: \url{https://github.com/FarjadMalik/aigoeswild}.}.
First we trained the baseline CNN on the full dataset $\mathcal{D}_{orig}$.
Next, we trained 4 models with different instantiations of $\mathcal{D}_{bal}$ for phase 1 of two-phase training. 
\begin{enumerate}
    \item $\mathcal{D}_{bal}^1$: ROS (oversampling) classes with less than 5k images until 5k, see appendix fig. \ref{fig:ROS}.
    \item $\mathcal{D}_{bal}^2$: RUS (undersampling) classes with more than 15k images until 15k.
    \item $\mathcal{D}_{bal}^3$: ROS classes with less than 5k images until 5k as in 1., and RUS classes with more than 15k images until 15k as in 2. Shown in fig. \ref{fig:ROS&RUS} in the appendix.
    \item $\mathcal{D}_{bal}^4$: ROS classes with less than 5k images until 5k as in 1., and RUS classes with more than 5k images until 5k.
\end{enumerate}

We used a lower sample ratio for classes with very few images to avoid overfitting (appendix \ref{app:Sampling}).
As evaluation metrics we used not only top-1 accuracy but also precision, recall and F1-score, since these metrics are more informative to class-imbalance. 
We report their values macro-averaged over classes as well as the class specific values (in appendix tables \ref{tab:baseline1}-\ref{tab:rus_perspecies_stat}).
The results of the models after phase 1 correspond to the results that we would obtain by only using ROS, RUS or a combination of both 
(and no two-phase training). These results will thus serve as a baseline.

\section{Results} \label{sec:results}
\begin{table}
    \footnotesize
    \centering
    \begin{tabular}{lrrrr}
    \toprule
        \textbf{Model} & \textbf{Phase 1: Accuracy} & \textbf{Phase 2: Accuracy} & \textbf{Phase 1: F1} & \textbf{Phase 2: F1} \\ \midrule
        $\mathcal{D}_{orig}$: Baseline & \textbf{0.8527} & / & {0.3944} & / \\
        $\mathcal{D}_{bal}^1$: {ROS} & {0.8326} & \textbf{0.8528} & {0.3843} & {0.4012}\\
        $\mathcal{D}_{bal}^2$: {RUS} & {0.8012} & {0.8491} & {0.3681} & \textbf{0.4147} \\
        $\mathcal{D}_{bal}^3$: ROS\&RUS(15K) & 0.8346 & 0.8454  & \textbf{0.4179} & 0.4094\\
        $\mathcal{D}_{bal}^4$: ROS\&RUS(5K) & 0.7335 & 0.8066 & 0.3620 & 0.4001\\ \bottomrule
    \end{tabular}
    \caption{Model Comparison - Top-1 accuracy and Macro F1-score.}
    \label{tab:modelcomparison_top1_F1}
\end{table}
\paragraph{Accuracy and Macro F1.} Table \ref{tab:modelcomparison_top1_F1} shows the accuracy and F1-score of the models after the 1st and the 2nd phase\footnote{Appendix \ref{app:extra-results} contains more results and in-depth discussion.}.
Training on more balanced datasets reduces accuracy in phase 1 for all models compared to the baseline which was trained on the imbalanced dataset $\mathcal{D}_{orig}$. 
However, further training the classification layers in phase 2 on the full dataset increases accuracy back to more or less the baseline level for all models (except ROS\&RUS(5K)), meaning that two-phase training lost little to no accuracy.
The phase 2 mean accuracy is substantially higher than the phase 1 mean accuracy.

The F1-scores of most models 
also drop in phase 1.
Interestingly, phase 2 raises the F1-score of most models 
again, and all models obtain an F1-score after phase 2 that is higher than the baseline: 3.0\% on average.
The RUS model obtains the highest F1-score after phase 2: an increase of 5.1\% compared to the baseline, while the ROS\&RUS(15K) model obtain the highest F1-score overall\footnote{We consider the F1-score of ROS\&RUS(15K) after phase 1 an anomaly which needs further analysis.}.
Most two-phase trained models outperform their counterparts which were only trained on more balanced datasets.
As for the accuracy, the mean F1-score in phase 2 is substantially higher than the mean F1-score in phase 1: 6.1\%.

These observations lead us to conclude that 1) two-phase training outperforms using only sampling techniques across most sampling regimes, and 2) two-phase training can increase the F1-score without substantial loss in accuracy, meaning it improves minority class predictions with very limited loss in majority class performance.
These findings are in line with the results of \citep{lee2016plankton}, though they report greater increases in F1-scores than us, possibly due to an even more imbalanced dataset.
They also find RUS to work best for creating $\mathcal{D}_{bal}$ for phase 1.
The F1-scores are substantially lower than the accuracies
(idem for precision and recall, appendix tables \ref{tab:modelcomparison_precision}-\ref{tab:modelcomparison_recall}). 
This is because the class-specific values for these metrics are high for the majority classes, but extremely low for many minority classes, confirming that the imbalanced data creates a bias towards the majority classes.

\begin{figure}
    \centering
    \begin{subfigure}{0.5\textwidth}
        \centering
          \includegraphics[width=0.96\linewidth,keepaspectratio]{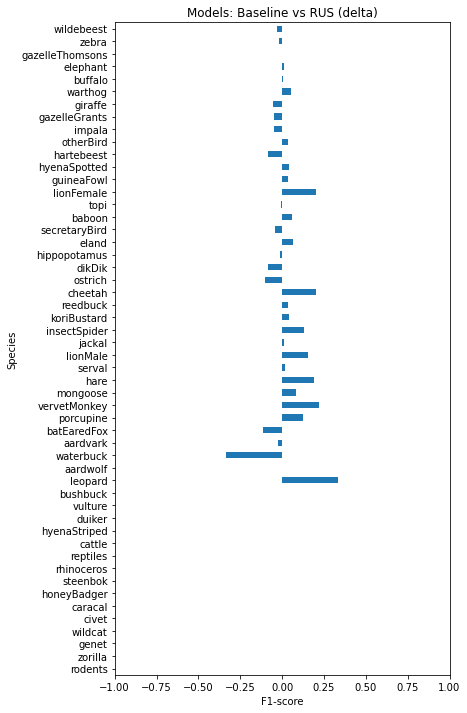}
        \caption{}
          \label{fig:f1_comparison_delta_base_rus}
    \end{subfigure}%
    \begin{subfigure}{0.5\textwidth}
        \centering
        \includegraphics[width=0.96\linewidth,keepaspectratio]{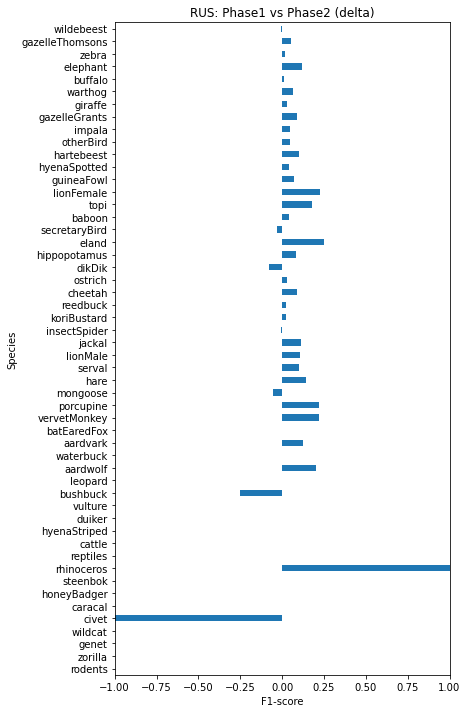}
        \caption{}
          \label{fig:f1_comparison_delta_rus_phases}
    \end{subfigure}
    \caption{Relative difference in F1-score per species of (\subref{fig:f1_comparison_delta_base_rus}) the two-phase RUS model vs. the baseline, and (\subref{fig:f1_comparison_delta_rus_phases}) phase 2 vs. phase 1 of the RUS-model. The appendix contains larger versions: figs. \ref{fig:f1_comparison_delta_base_rus_l}, \ref{fig:f1_comparison_delta_rus_phases_l}. Species are sorted in descending order according to their occurrence frequency.}
    \label{fig:f1_comparisons}
\end{figure}

\paragraph{Class-specific performance.} 
Class-specific F1-scores increase with two-phase training for the majority of (minority) classes.
Two-phase training with RUS leads to the greatest average increase of F1-score per class: 3\% (ignoring the classes for which the F1-score remained 0.0\%).
This increase is most notable for minority classes.
RUS performing best is remarkable, since we trained the RUS model in phase 1 with only 85k images, compared to 131k--231k for the other models.
Fig. \ref{fig:f1_comparison_delta_base_rus} shows the changes in F1-score due to two-phase training with RUS.

\section{Conclusion}
We explored the use of two-phase training to mitigate the class imbalance issue for camera trap image classification with CNNs.
We conclude that 1) two-phase training outperforms using only sampling techniques across most sampling regimes, and 2) two-phase training improves minority class predictions with very limited loss in majority class performance, compared to training on the imbalanced dataset only.
In the future we would like to rerun our experiments with different random seeds to obtain more statistically convincing results, compare two-phase training to other algorithm-level imbalance mitigation techniques, and test it on varying dataset sizes and levels of class imbalance.

\clearpage
\bibliography{main}

\clearpage
\appendix
\section{Appendix}

\subsection{Figures}
\begin{figure}[h]
\includegraphics[width=8cm,keepaspectratio]{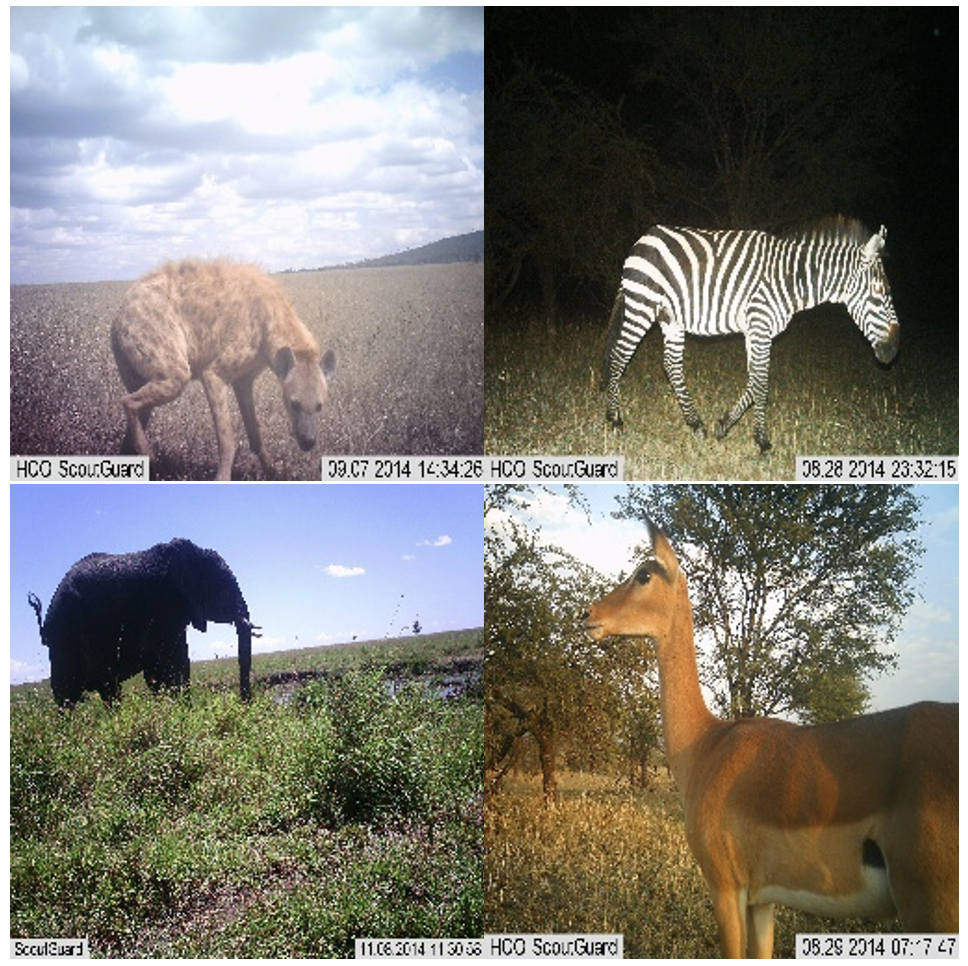}
\centering
\caption{Four examples of camera trap images.}
\label{fig:images}
\end{figure}

\begin{figure}
\includegraphics[width=11cm,keepaspectratio]{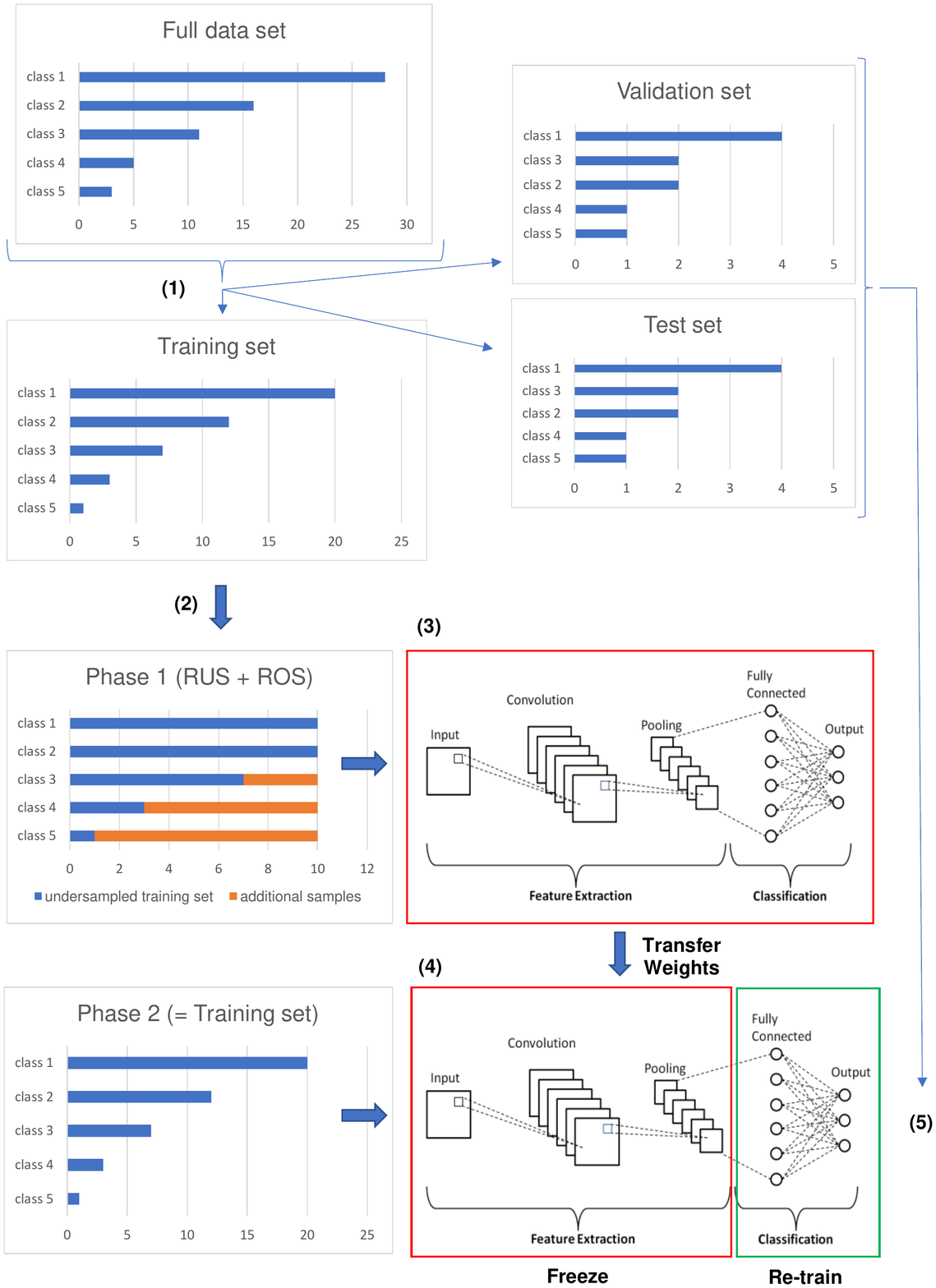}
\centering
\caption{Schematic overview of a general two-phase training implementation. }
\label{fig:two-phase fig}
\end{figure}

\begin{figure}
\includegraphics[width=12cm,keepaspectratio]{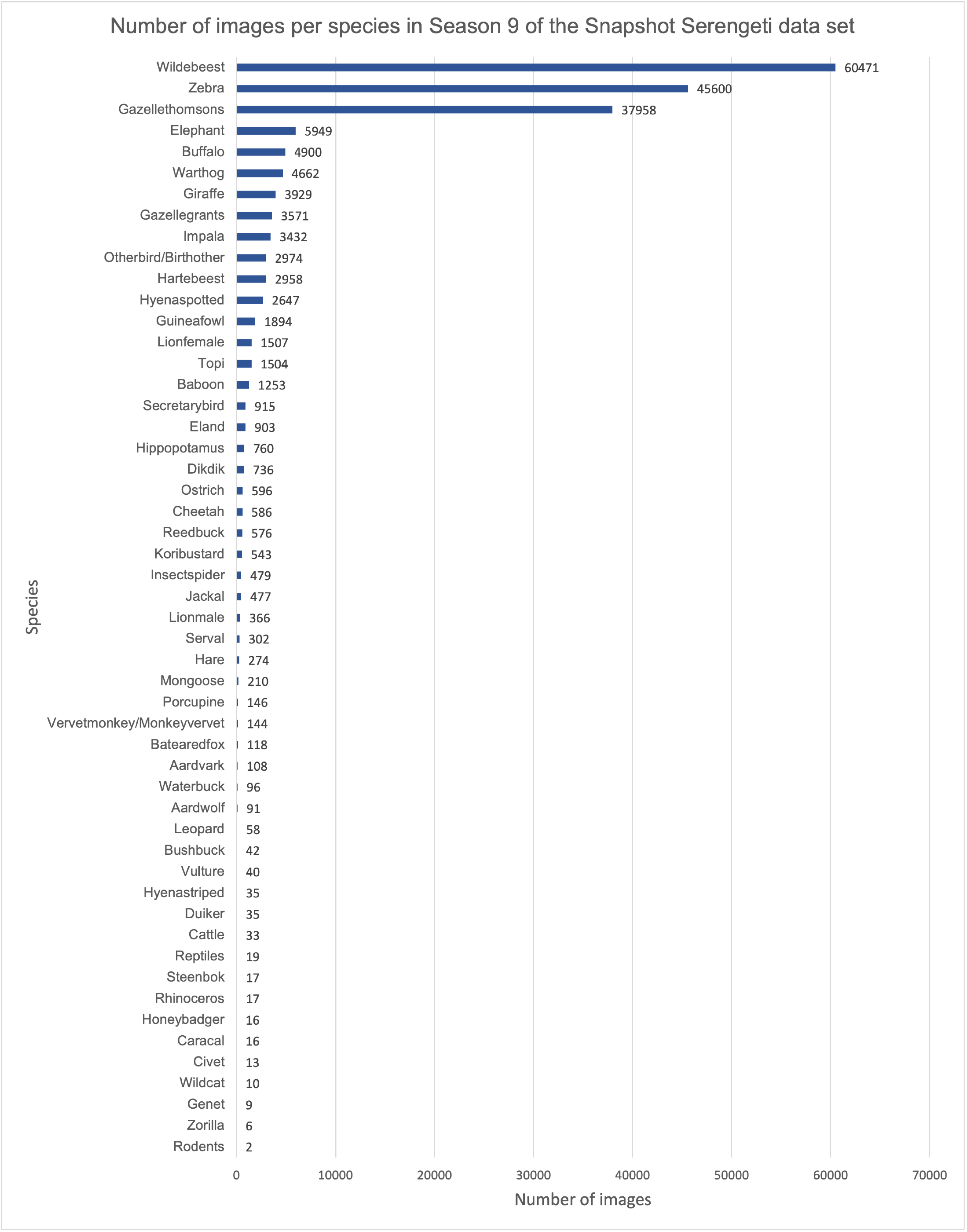}
\centering
\caption{Class-distribution of the 9th season of the SS dataset.}
\label{fig:season9_species fig}
\end{figure}

\begin{figure}
\includegraphics[width=12cm,keepaspectratio]{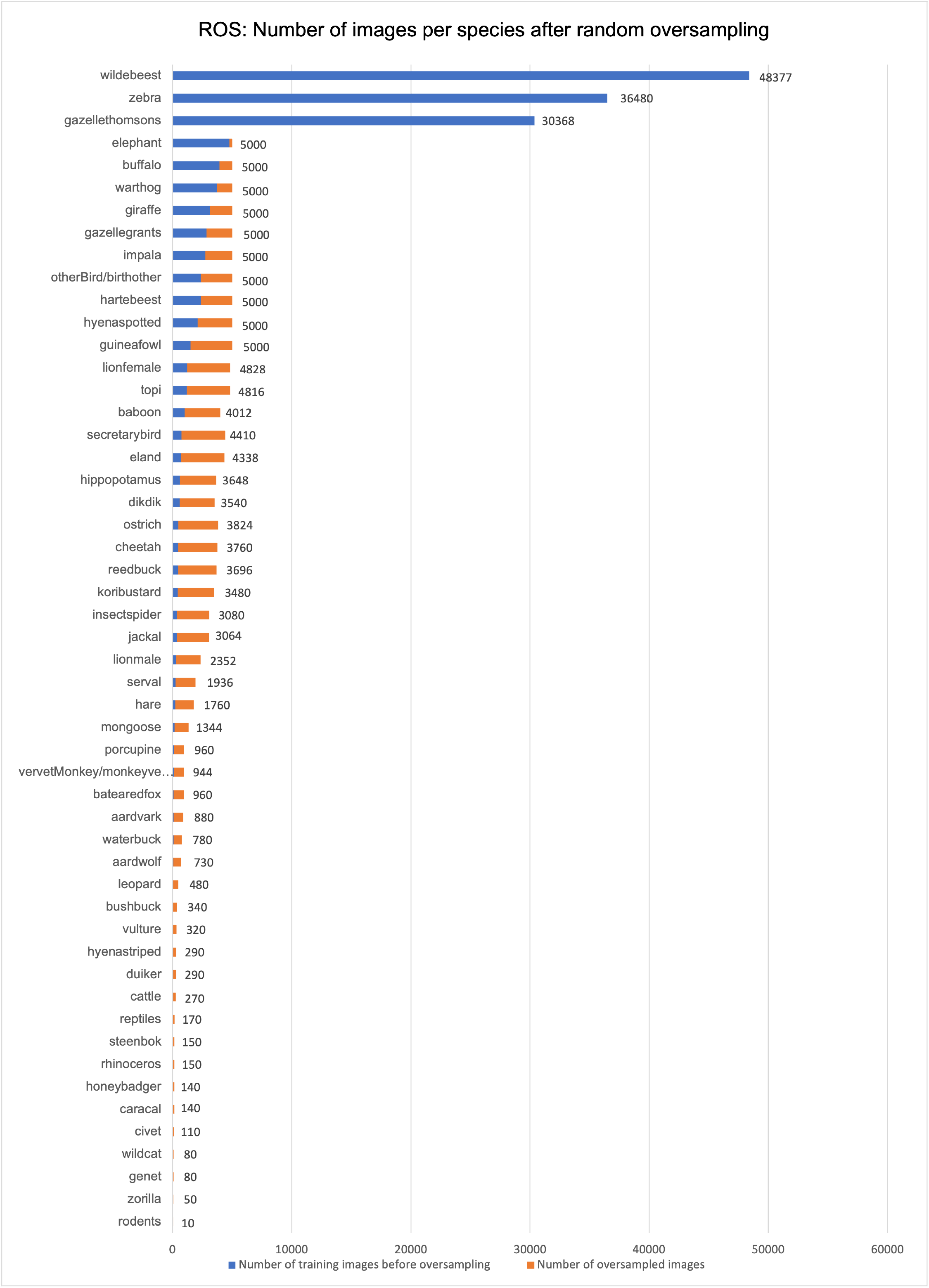}
\centering
\caption{Class-distribution of the dataset used in the first phase for the ROS model.}
\label{fig:ROS}
\end{figure}

\begin{figure}
\includegraphics[width=12cm,keepaspectratio]{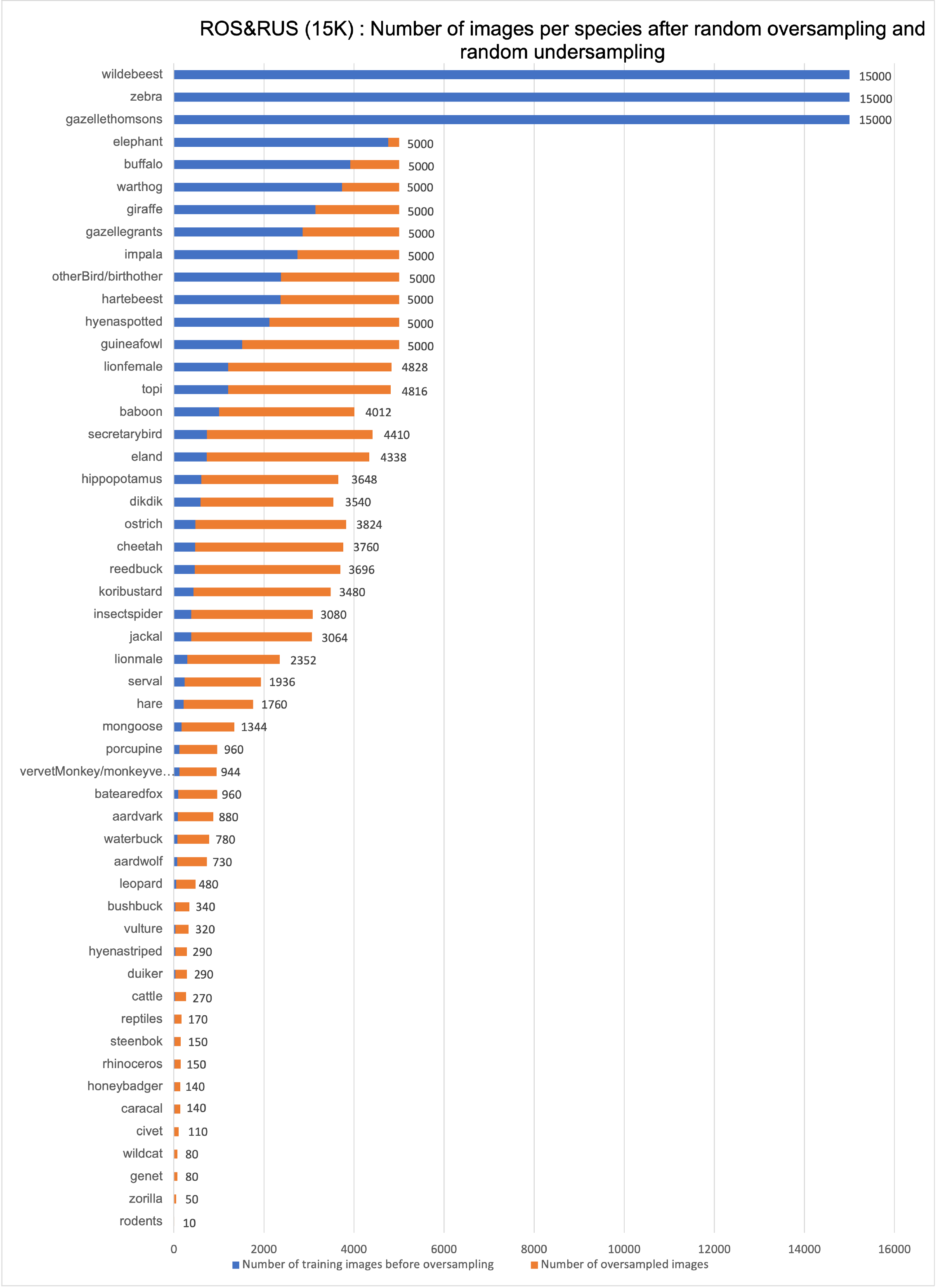}
\centering
\caption{Class-distribution of the dataset used in the first phase for the ROS\&RUS(15K) model.}
\label{fig:ROS&RUS}
\end{figure}

\subsection{Experiments \& hyperparameters} \label{app:hyperparams}

We omitted images that contain more than one species, images that do not contain any animal (class `blank'), as well as images belonging to the class `human'.
We split our dataset following a 80\%-10\%-10\% train, validation and test split. We chose for the ResNet-18 architecture \citep{he2016deep}, which performed well in previous camera trap literature \citep{norouzzadeh2018automatically,tabak2019machine,willi2019identifying}. We used the ADAM optimizer, a batch-size of 64 and a learning rate of 0.001. Data augmentation and early stopping (for the baseline as well as for both phases of all 2-phase models) were used to avoid overfitting. 
We used categorical cross-entropy as loss function.
We trained the baseline model for 10 epochs on the training set, and the ROS model, the RUS model, the ROS\&RUS(15K) model and the ROS\&RUS(5K) model in the first phase for respectively 10, 14, 14 and 17 epochs.
In the second phase, they were trained for an additional 7, 14, 8 and 14 epochs.
Hyperparameters, incl. the ROS and RUS thresholds of 5k and 15k, were taken from existing literature, or experimentally tuned using a grid search.

After training all the four models in the first phase, we extracted the weights of all the 18 layers of the models. 
We loaded these weights into four new models with the same ResNet-18 architecture. Before fine-tuning these models on the original, imbalanced, training data set, we froze the weights in the 17 convolutional layers. Only the single fully connected layer of the ResNet-18 model was fine-tuned on the training set. By freezing the majority of the layers, the number of trainable parameters was reduced from slightly over 11 million to only 26,676. This implies that training the second phase is far less computationally demanding than training the models in the first phase. 

\subsection{Sampling} 
\label{app:Sampling}

\subsubsection{ROS}
For the ROS data set, we oversampled all the classes that contained less than 5000 images to a maximum of 5000 images. The three majority classes (see fig. \ref{fig:season9_species fig}) were not oversampled. Since our minority classes contained very little samples, we did not oversample these classes up to 5000 images, as this would likely result in overfitting.
Instead, for all classes that contained less than 100 images, we applied a sample ratio of 10. This means that the image(s) for a certain class were oversampled until 10 times the original number of images was present in the data set. This is similar to \citet{schneider2020three}, who supplemented the minority classes with less than 100 images with fixed augmentations until at least 100 images were available. For their data set, this also implied a sampling ratio of roughly 10 for the smallest classes.

In order to obtain a more balanced dataset, this sampling ratio was gradually decreased. For classes that originally contained between 100 \& 500, 500 \& 1000, and 1000 \& 5000, we used sample ratios of respectively 8, 6 and 4. We thresholded oversampling at 5000 for the classes that contained less than 5000 images. This threshold was set experimentally and is higher than the threshold of 1000 that \citet{lee2016plankton} used when trying two-phase training in combination with ROS. We found that oversampling to 1000 images was not sufficient for our data set. It is, however, to be noted that a threshold of 5000 implies increased computational requirements. While the original training set contained slightly over 155,000 images, the ROS data set contained slightly over 231,000 images. 

\subsubsection{RUS}
To avoid an increase in the number of samples in our dataset, such as in the case of ROS, we decided to also train a model on a randomly undersampled dataset. 
\citet{lee2016plankton} obtained the best results for two-phase training when using RUS only. The authors put their threshold at 5000, meaning that all the classes with more than 5000 images were undersampled until they held at most 5000 samples. Experimental results indicated that this threshold was too low for us when it was not used in combination with ROS. The vast reduction in the size of the data set might be a possible explanation for this. Therefore, we set the undersampling threshold at 15,000 for the RUS model. The RUS data set thus looks exactly the same as the original training data set, except for the fact that the three majority classes all contain 15,000 images instead of respectively 48,377; 36,480; and 30,368 images. This brought the total number of images for the RUS data set to slightly over 85,000. 

\subsubsection{ROS\&RUS(15K) and ROS\&RUS(5K) }

The limited previous literature on two-phase training has not used this method in combination with both ROS and RUS at the same time in the first phase. 
Therefore, we decided to explore the effectiveness of two-phase training when both ROS and RUS are used in the first phase. For this purpose, we created two more balanced data sets, which we refer to as ROS\&RUS(15K) and ROS\&RUS(5K). 
We obtained the ROS\&RUS(15K) data set by combining the over- and undersampling regimes of the ROS model and the RUS model respectively.
The class-distribution of this data set is depicted in figure \ref{fig:ROS&RUS}. Interestingly, by using these two sampling regimes, we got a data set that only contains 6000 images more than the original training set.

\subsection{Extra results \& discussion} \label{app:extra-results}

\subsubsection{Overall performance of the models: Overall top-1 accuracy and \& Macro F1-score}

Table \ref{tab:modelcomparison_top1_F1} showed the top-1 accuracy and the Macro F1-score of the models for the 1st and the 2nd phase. 
Tables \ref{tab:modelcomparison_precision} and \ref{tab:modelcomparison_recall} additionally show the precision and recall.

\begin{table}
    \centering
    \begin{tabular}{lrrr}\toprule
        \textbf{Model} & \textbf{First Phase}  & \textbf{Second Phase} \\ \midrule
        Baseline         & 0.5055      & N.A.       \\
        \textbf{ROS}    & \textbf{0.4691}      & \textbf{0.5053}     \\
        \textbf{RUS}    & \textbf{0.4609}      & \textbf{0.5319}     \\
        ROS\&RUS   (15K) & 0.4519      & 0.5284     \\
        ROS\&RUS   (5K)  & 0.3702      & 0.5640     \\ \bottomrule
    \end{tabular}
    \caption{Model Comparison - Precision.}
    \label{tab:modelcomparison_precision}
\end{table}

\begin{table}
    \centering
    \begin{tabular}{lrrr} \toprule
        \textbf{Model} & \textbf{First Phase}  & \textbf{Second Phase} \\ \midrule
        Baseline         & 0.3558      & N.A.       \\
        \textbf{ROS}              & \textbf{0.3689}      & \textbf{0.3635}     \\
        \textbf{RUS}              & \textbf{0.3592}      & \textbf{0.3648}     \\
        ROS\&RUS   (15K) & 0.4094      & 0.3563     \\
        ROS\&RUS   (5K)  & 0.3974      & 0.3438      \\ \bottomrule
    \end{tabular}
    \caption{Model Comparison - Recall.}
    \label{tab:modelcomparison_recall}
\end{table}

The values of the class-weighted F1-score (same for precision and recall as shown in appendix table \ref{tab:modelcomparison_precision} and \ref{tab:modelcomparison_recall}) are substantially lower than our overall accuracy. 
This is caused by the fact that the class-specific values for these metrics are very high for the majority classes, but extremely low for many minority classes. 
For example, the recall varies between 0\% for the minority classes and 96.0\% for the majority class. 
Many of these minority classes contain only few images in the training and the test set, which explains why the model performs poorly on them. 
This indicates that the imbalanced data indeed created a bias towards the majority class(es). 
This bias is also reflected in the difference in discrepancy between precision and recall for classes of different sizes. 
For the majority classes, we find a higher recall than precision. 
This indicates that the model predicts the majority class more often than necessary. 
For the minority class, the precision tends to be higher than the recall. 
This indicates that the baseline model could likely improve performance when more attention is diverted from the majority classes to the minority classes.

Taking both overall accuracy and macro F1-score into account, we can conclude that two-phase training can result in a small increase in F1-score without losing overall accuracy. 
This slight increase in F1-score can be attributed to a slight increase of both the precision and recall. 
The ROS and the RUS models seem to perform best, given that they improve upon the F1-score of the baseline model, without or with little decrease in overall accuracy. 
Generally, two-phase training also outperforms only using ROS, RUS or a combination of both. 
Training in two phases leads to an improvement in both overall accuracy and F1-score compared to the models that were trained in one phase on a more balanced dataset. 
An exception to this observation is the ROS\&RUS(15K) model. 
This model achieves the highest F1-score but performs slightly worse than the baseline model in terms of overall accuracy. 
From our results, we can see that the RUS model benefits more in terms of increase in precision when fine-tuning the classification layer. 
Possibly, training the ROS model for both phases with full majority classes limited the increase in precision for this model, since a focus on majority classes tends to lead to many false positives and thus a lower precision.

Our results are in accordance with findings of \citet{lee2016plankton}, who also concluded that two-phase training in combination with RUS leads to the highest increase in F1-score with respect to the baseline model. 
However, they managed to almost double their baseline F1-score from 17.73\% to 33.39\%. 
One possible explanation for this larger increase in F1-score might be that their dataset was more imbalanced than ours. 
This would mean that two-phase training becomes more effective when datasets become more imbalanced. 
Nevertheless, this thesis thus presents some evidence for the argument by \citet{lee2016plankton}, who stated that by making the dataset more balanced in the first phase, animal population information is lost and that this information consequently needs to be restored by fine-tuning the classification layer on the training set representing the population distribution of the ecosystem.

\subsubsection{Class-specific performance}

\begin{figure}
  \includegraphics[width=12.5cm,keepaspectratio]{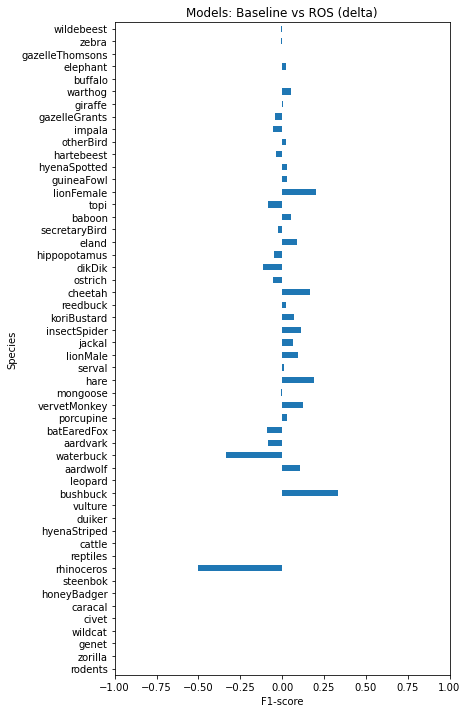}
  \caption{Percentage increase or decrease in F1-score per species of the ROS model compared to the baseline model.}
  \label{fig:f1_comparison_delta_base_ros_l}
\end{figure}

\begin{figure} %
  \includegraphics[width=12.5cm,keepaspectratio]{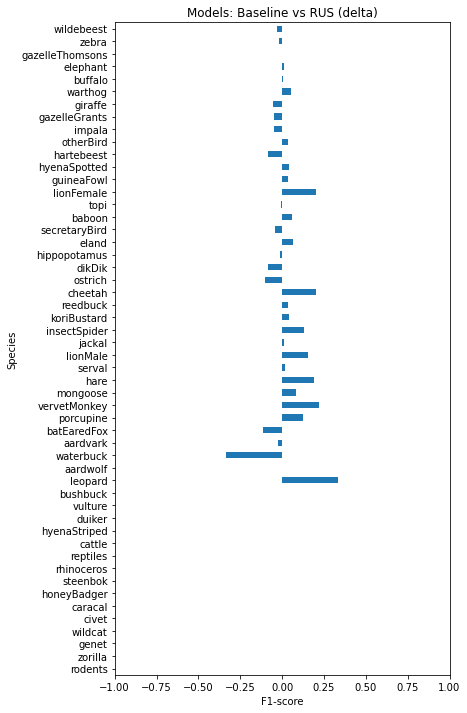}
  \caption{Percentage increase or decrease in F1-score per species of the RUS model compared to the baseline model.}
  \label{fig:f1_comparison_delta_base_rus_l}
\end{figure}

Table \ref{tab:ros_perspecies_stat} and table \ref{tab:rus_perspecies_stat} show respectively the class-specific results for the ROS and the RUS model, when trained using two-phase training. 
These tables follow the same structure as table \ref{tab:baseline1}, which shows the class-specific results for the baseline model, except for the fact that the Count(Train) column now depicts the number of samples that were used for training in the first phase only. In the second phase, all the four models were fine-tuned on the original training set. 
Figs. \ref{fig:f1_comparison_delta_base_ros_l} and \ref{fig:f1_comparison_delta_base_rus_l} show the relative change in F1-score when using ROS and RUS, respectively, versus the baseline.


\subsubsection{Discussion of two-phase training: phase 1 compared to phase 2}

\begin{figure}
  \includegraphics[width=12.5cm,keepaspectratio]{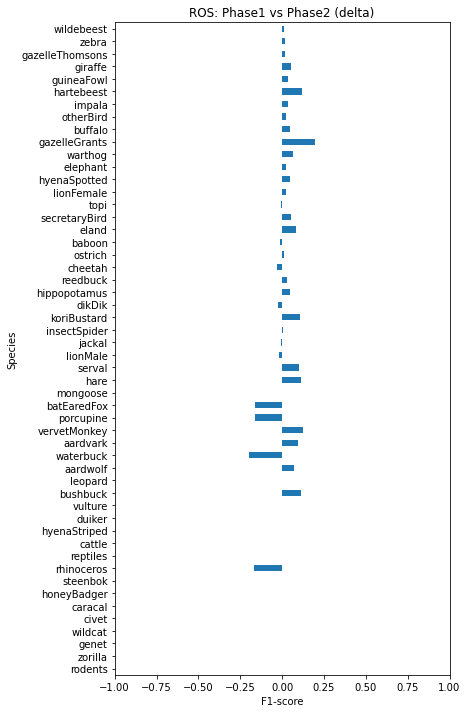}
  \caption{Percentage increase or decrease in F1-score per species of the second phase of the ROS model compared to the first phase.}
  \label{fig:f1_comparison_delta_ros_phases_l}
\end{figure}

\begin{figure}
  \includegraphics[width=12.5cm,keepaspectratio]{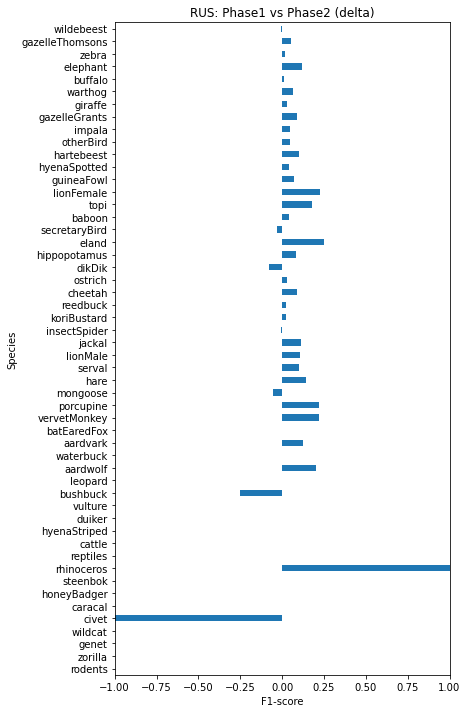}
  \caption{Percentage increase or decrease in F1-score per species of the second phase of the RUS model compared to the first phase.}
  \label{fig:f1_comparison_delta_rus_phases_l}
\end{figure}

Figs. \ref{fig:f1_comparison_delta_ros_phases_l} and \ref{fig:f1_comparison_delta_rus_phases_l} show the relative change in F1-score when using ROS and RUS, respectively, in phase 2 versus phase 1.

The large performance differences between the first and the second phase of the ROS, the RUS and the ROS\&RUS(5K) model can mainly be attributed to the vast increase in precision that is obtained when fine-tuning the classification layer. 
A possible explanation for this observation is that in the first phase, precision and recall are low given that the distribution of the training set does not match that one of the test set. 
At this stage, the model might be good at feature extraction for the minority class images. 
However, the classification layer is still not used to the real data distribution. 
Therefore, the models might predict the oversampled classes too often, leading to a lower precision. 
After fine-tuning the classification layer, the model might be better able to combine its enhanced feature extraction skills with respect to the minority class images with the knowledge it has on the real data distribution. 


\begin{table}
{%
\fontsize{9.0pt}{10.0pt}\selectfont
\centering
\begin{tabular}{lrrrrrr}
\toprule
\textbf{Species} & \textbf{Precision} & \textbf{Recall} & \textbf{F1-score} & \textbf{Count (Train)} & \textbf{Count (Test)} \\ \midrule
wildebeest & 0.8612 & 0.9601 & 0.908 & 48377 & 6047\\
zebra & 0.9096 & 0.9377 & 0.9234 & 36480 & 4560 \\
gazelleThomsons & 0.9116 & 0.9078 & 0.9097 & 30368 & 3795 \\
elephant & 0.9382 & 0.6902 & 0.7953 & 4761 & 594 \\
buffalo & 0.804 & 0.6531 & 0.7207 & 3920 & 490 \\
warthog & 0.5548 & 0.6631 & 0.6041 & 3730 & 466 \\
giraffe & 0.9286 & 0.7628 & 0.8375 & 3145 & 392 \\
gazelleGrants & 0.5641 & 0.4314 & 0.4889 & 2857 & 357 \\
impala & 0.7595 & 0.7551 & 0.7573 & 2746 & 343 \\
otherBird & 0.5246 & 0.431 & 0.4732 & 2380 & 297 \\
hartebeest & 0.685 & 0.6339 & 0.6585 & 2368 & 295 \\
hyenaSpotted & 0.5202 & 0.6326 & 0.5709 & 2119 & 264 \\
guineaFowl & 0.7963 & 0.6825 & 0.735 & 1516 & 189 \\
lionFemale & 0.9545 & 0.28 & 0.433 & 1207 & 150 \\
topi & 0.5764 & 0.5533 & 0.5646 & 1204 & 150 \\
baboon & 0.7021 & 0.528 & 0.6027 & 1003 & 125 \\
secretaryBird & 0.9012 & 0.8111 & 0.8538 & 735 & 90 \\
eland & 0.9429 & 0.3667 & 0.528 & 723 & 90 \\
hippopotamus & 0.9143 & 0.8421 & 0.8767 & 608 & 76 \\
dikDik & 0.6747 & 0.7671 & 0.7179 & 590 & 73 \\
ostrich & 0.6667 & 0.339 & 0.4494 & 478 & 59 \\
cheetah & 0.4615 & 0.2069 & 0.2857 & 470 & 58 \\
reedbuck & 0.7292 & 0.614 & 0.6667 & 462 & 57 \\
koriBustard & 0.4677 & 0.537 & 0.5 & 435 & 54 \\
insectSpider & 0.3636 & 0.0851 & 0.1379 & 385 & 47 \\
jackal & 0.4583 & 0.234 & 0.3099 & 383 & 47 \\
lionMale & 0.375 & 0.1667 & 0.2308 & 294 & 36 \\
serval & 0.6 & 0.3 & 0.4 & 242 & 30 \\
hare & 0.9231 & 0.4444 & 0.6 & 220 & 27 \\
mongoose & 0.4 & 0.1905 & 0.2581 & 168 & 21 \\
vervetMonkey & 0.0 & 0.0 & 0.0 & 120 & 12 \\
porcupine & 1.0 & 0.2857 & 0.4444 & 118 & 14 \\
batEaredFox & 0.75 & 0.2727 & 0.4 & 96 & 11 \\
aardvark & 1.0 & 0.2 & 0.3333 & 88 & 10 \\
waterbuck & 0.6667 & 0.2222 & 0.3333 & 78 & 9 \\
aardwolf & 1.0 & 0.1111 & 0.2 & 73 & 9 \\
leopard & 0.0 & 0.0 & 0.0 & 48 & 5 \\
bushbuck & 0.0 & 0.0 & 0.0 & 34 & 4 \\
vulture & 0.0 & 0.0 & 0.0 & 32 & 4 \\
duiker & 0.0 & 0.0 & 0.0 & 29 & 3 \\
hyenaStriped & 0.0 & 0.0 & 0.0 & 29 & 3 \\
cattle & 0.0 & 0.0 & 0.0 & 27 & 3 \\
reptiles & 0.0 & 0.0 & 0.0 & 17 & 2 \\
rhinoceros & 1.0 & 1.0 & 1.0 & 15 & 1 \\
steenbok & 0.0 & 0.0 & 0.0 & 15 & 1 \\
honeyBadger & 0.0 & 0.0 & 0.0 & 14 & 1 \\
caracal & 0.0 & 0.0 & 0.0 & 14 & 1 \\
civet & 0.0 & 0.0 & 0.0 & 11 & 1 \\
wildcat & 0.0 & 0.0 & 0.0 & 8 & 1 \\
genet & 0.0 & 0.0 & 0.0 & 8 & 1 \\
zorilla & 0.0 & 0.0 & 0.0 & 5 & 1 \\
rodents & 0.0 & 0.0 & 0.0 & 1 & 1 \\
\midrule
\textbf{Macro} & \textbf{0.5055} & \textbf{0.3558} & \textbf{0.3944} & \textbf{155254} & \textbf{19377} \\ 
\bottomrule
\end{tabular}}%
\caption{Baseline Model - Per Species Statistics.}
\label{tab:baseline1}
\end{table}

\begin{table}
{%
\fontsize{9.0pt}{10.0pt}\selectfont
\centering
\begin{tabular}{lrrrrrr}
\toprule
\textbf{Species} & \textbf{Precision} & \textbf{Recall} & \textbf{F1-score} & \textbf{Count (Train)} & \textbf{Count (Test)} \\ \midrule
wildebeest & 0.8438 & 0.9666 & 0.901 & 48377 & 6047 \\
zebra & 0.9173 & 0.9143 & 0.9158 & 36480 & 4560 \\
gazelleThomsons & 0.8813 & 0.9333 & 0.9066 & 30368 & 3795 \\
giraffe & 0.9112 & 0.7857 & 0.8438 & 5000 & 392 \\
guineaFowl & 0.8506 & 0.6931 & 0.7638 & 5000 & 189 \\
hartebeest & 0.77 & 0.522 & 0.6222 & 5000 & 295 \\
impala & 0.7055 & 0.7055 & 0.7055 & 5000 & 343 \\
otherBird & 0.6058 & 0.4242 & 0.499 & 5000 & 297 \\
buffalo & 0.765 & 0.6776 & 0.7186 & 5000 & 490 \\
gazelleGrants & 0.5312 & 0.381 & 0.4437 & 5000 & 357 \\
warthog & 0.7205 & 0.603 & 0.6565 & 5000 & 466 \\
elephant & 0.8893 & 0.7576 & 0.8182 & 5000 & 594 \\
hyenaSpotted & 0.6368 & 0.5644 & 0.5984 & 5000 & 264 \\
lionFemale & 0.7368 & 0.56 & 0.6364 & 4828 & 150 \\
topi & 0.6552 & 0.38 & 0.481 & 4816 & 150 \\
secretaryBird & 0.9565 & 0.7333 & 0.8302 & 4410 & 90 \\
eland & 0.8776 & 0.4778 & 0.6187 & 4338 & 90 \\
baboon & 0.8481 & 0.536 & 0.6569 & 4012 & 125 \\
ostrich & 0.6296 & 0.2881 & 0.3953 & 3824 & 59 \\
cheetah & 0.6452 & 0.3448 & 0.4494 & 3760 & 58 \\
reedbuck & 0.8889 & 0.5614 & 0.6882 & 3696 & 57 \\
hippopotamus & 0.8592 & 0.8026 & 0.8299 & 3648 & 76 \\
dikDik & 0.5435 & 0.6849 & 0.6061 & 3540 & 73 \\
koriBustard & 0.7353 & 0.463 & 0.5682 & 3480 & 54 \\
insectSpider & 0.375 & 0.1915 & 0.2535 & 3080 & 47 \\
jackal & 0.4545 & 0.3191 & 0.375 & 3064 & 47 \\
lionMale & 0.6154 & 0.2222 & 0.3265 & 2352 & 36 \\
serval & 0.8889 & 0.2667 & 0.4103 & 1936 & 30 \\
hare & 0.9048 & 0.7037 & 0.7917 & 1760 & 27 \\
mongoose & 1.0 & 0.1429 & 0.25 & 1344 & 21 \\
batEaredFox & 1.0 & 0.1818 & 0.3077 & 960 & 11 \\
porcupine & 0.7143 & 0.3571 & 0.4762 & 960 & 14 \\
vervetMonkey & 0.25 & 0.0833 & 0.125 & 944 & 12 \\
aardvark & 0.3333 & 0.2 & 0.25 & 880 & 10 \\
waterbuck & 0.0 & 0.0 & 0.0 & 780 & 9 \\
aardwolf & 0.5 & 0.2222 & 0.3077 & 730 & 9 \\
leopard & 0.0 & 0.0 & 0.0 & 480 & 5 \\
bushbuck & 0.5 & 0.25 & 0.3333 & 340 & 4 \\
vulture & 0.0 & 0.0 & 0.0 & 320 & 4 \\
duiker & 0.0 & 0.0 & 0.0 & 290 & 3 \\
hyenaStriped & 0.0 & 0.0 & 0.0 & 290 & 3 \\
cattle & 0.0 & 0.0 & 0.0 & 270 & 3 \\
reptiles & 0.0 & 0.0 & 0.0 & 170 & 2 \\
rhinoceros & 0.3333 & 1.0 & 0.5 & 150 & 1 \\
steenbok & 0.0 & 0.0 & 0.0 & 150 & 1 \\
honeyBadger & 0.0 & 0.0 & 0.0 & 140 & 1 \\
caracal & 0.0 & 0.0 & 0.0 & 140 & 1 \\
civet & 0.0 & 0.0 & 0.0 & 110 & 1 \\
wildcat & 0.0 & 0.0 & 0.0 & 80 & 1 \\
genet & 0.0 & 0.0 & 0.0 & 80 & 1 \\
zorilla & 0.0 & 0.0 & 0.0 & 50 & 1 \\
rodents & 0.0 & 0.0 & 0.0 & 10 & 1 \\
\midrule
\textbf{Macro} & \textbf{0.5053} & \textbf{0.3635} & \textbf{0.4012} & \textbf{231437} & \textbf{19377} \\ 
\bottomrule
\end{tabular}}
\caption{ROS Model (two-phase training) - Per Species Statistics.}
\label{tab:ros_perspecies_stat}
\end{table}

\begin{table}
{%
\fontsize{9.0pt}{10.0pt}\selectfont
\centering
\begin{tabular}{lrrrrrr}
\toprule
\textbf{Species} & \textbf{Precision} & \textbf{Recall} & \textbf{F1-score} & \textbf{Count (Train)} & \textbf{Count (Test)} \\ \midrule
wildebeest & 0.8099 & 0.9558 & 0.8768 & 15000 & 6047 \\
gazelleThomsons & 0.91 & 0.9086 & 0.9093 & 15000 & 3795 \\
zebra & 0.8951 & 0.9154 & 0.9051 & 15000 & 4560 \\
elephant & 0.8598 & 0.7643 & 0.8093 & 4761 & 594 \\
buffalo & 0.8478 & 0.6367 & 0.7273 & 3920 & 490 \\
warthog & 0.7104 & 0.6159 & 0.6598 & 3730 & 466 \\
giraffe & 0.9144 & 0.6811 & 0.7807 & 3145 & 392 \\
gazelleGrants & 0.6383 & 0.3361 & 0.4404 & 2857 & 357 \\
impala & 0.727 & 0.691 & 0.7085 & 2746 & 343 \\
otherBird & 0.6703 & 0.4108 & 0.5094 & 2380 & 297 \\
hartebeest & 0.586 & 0.5661 & 0.5759 & 2368 & 295 \\
hyenaSpotted & 0.6622 & 0.5644 & 0.6094 & 2119 & 264 \\
guineaFowl & 0.8035 & 0.7354 & 0.768 & 1516 & 189 \\
lionFemale & 0.7265 & 0.5667 & 0.6367 & 1207 & 150 \\
topi & 0.7528 & 0.4467 & 0.5607 & 1204 & 150 \\
baboon & 0.8701 & 0.536 & 0.6634 & 1003 & 125 \\
secretaryBird & 0.9286 & 0.7222 & 0.8125 & 735 & 90 \\
eland & 0.9286 & 0.4333 & 0.5909 & 723 & 90 \\
hippopotamus & 0.8784 & 0.8553 & 0.8667 & 608 & 76 \\
dikDik & 0.6825 & 0.589 & 0.6324 & 590 & 73 \\
ostrich & 0.6364 & 0.2373 & 0.3457 & 478 & 59 \\
cheetah & 0.8333 & 0.3448 & 0.4878 & 470 & 58 \\
reedbuck & 0.85 & 0.5965 & 0.701 & 462 & 57 \\
koriBustard & 0.5472 & 0.537 & 0.5421 & 435 & 54 \\
insectSpider & 0.3143 & 0.234 & 0.2683 & 385 & 47 \\
jackal & 0.5 & 0.234 & 0.3188 & 383 & 47 \\
lionMale & 0.625 & 0.2778 & 0.3846 & 294 & 36 \\
serval & 0.6923 & 0.3 & 0.4186 & 242 & 30 \\
hare & 0.9048 & 0.7037 & 0.7917 & 220 & 27 \\
mongoose & 0.4286 & 0.2857 & 0.3429 & 168 & 21 \\
porcupine & 0.8571 & 0.4286 & 0.5714 & 120 & 14 \\
vervetMonkey & 0.3333 & 0.1667 & 0.2222 & 118 & 12 \\
batEaredFox & 0.6667 & 0.1818 & 0.2857 & 96 & 11 \\
aardvark & 0.6667 & 0.2 & 0.3077 & 88 & 10 \\
waterbuck & 0.0 & 0.0 & 0.0 & 78 & 9 \\
aardwolf & 1.0 & 0.1111 & 0.2 & 73 & 9 \\
leopard & 1.0 & 0.2 & 0.3333 & 48 & 5 \\
bushbuck & 0.0 & 0.0 & 0.0 & 34 & 4 \\
vulture & 0.0 & 0.0 & 0.0 & 32 & 4 \\
duiker & 0.0 & 0.0 & 0.0 & 29 & 3 \\
hyenaStriped & 0.0 & 0.0 & 0.0 & 29 & 3 \\
cattle & 0.0 & 0.0 & 0.0 & 27 & 3 \\
reptiles & 0.0 & 0.0 & 0.0 & 17 & 2 \\
rhinoceros & 1.0 & 1.0 & 1.0 & 15 & 1 \\
steenbok & 0.0 & 0.0 & 0.0 & 15 & 1 \\
honeyBadger & 0.0 & 0.0 & 0.0 & 14 & 1 \\
caracal & 0.0 & 0.0 & 0.0 & 14 & 1 \\
civet & 0.0 & 0.0 & 0.0 & 11 & 1 \\
wildcat & 0.0 & 0.0 & 0.0 & 8 & 1 \\
genet & 0.0 & 0.0 & 0.0 & 8 & 1 \\
zorilla & 0.0 & 0.0 & 0.0 & 5 & 1 \\
rodents & 0.0 & 0.0 & 0.0 & 1 & 1 \\
\midrule
\textbf{Macro} & \textbf{0.5319} & \textbf{0.3648} & \textbf{0.4171} & \textbf{85029} & \textbf{19377} \\ 
\bottomrule
\end{tabular}}
\caption{RUS Model (two-phase training) - Per Species Statistics.}
\label{tab:rus_perspecies_stat}
\end{table}

\subsection{Limitations \& future work}

We did not achieve a similar overall accuracy for our baseline model compared to most other works. 
Although achieving a high overall accuracy was not the main goal of this thesis, we consider our baseline performance as a limitation since it remains unknown whether two-phase training could also lead to increase in F1-score without losing overall accuracy, when the accuracy is very high. 

Second, the results that we reported at the start of this chapter need to be interpreted with care. In this paper, we aimed to treat each class equally. We did this by weighing our class-specific performance metrics by class rather than by sample and by not excluding any classes, regardless of their size. As a result, we obtained performance metrics that were more informative than the overall accuracy, which is generally biased towards the majority class. 
However, given that there were many minority classes with very few images in the training and the test set, high increases and decreases in class-specific performances were sometimes observed, though they only represent predictions on a few images. 
The performance for these extremely small classes mostly remained unchanged over the different models, meaning that their influence on the results is limited. 
Nevertheless, these classes do contribute to the values for the macro performance metrics and drive the differences of these values between the models slightly towards zero.
Therefore, both the macro performance metrics as well as the class-specific performance metrics need to be interpreted cautiously.

Third, the above explained limitation could be partly mitigated by training the different models several times, with a different random initialisation of the weights. This would allow us to report average performances for the different models, which would make our results more robust. However, the large size of the dataset implied long training times and prevented us from pursuing this preferred strategy.

\end{document}